\documentclass[11pt]{article}
\usepackage{colacl}
\usepackage{rotate}

\title{Exploiting auxiliary distributions
      in stochastic unification-based grammars}
\author{Mark Johnson\thanks{This research was supported by
        NSF awards 9720368, 9870676 and 9812169.}  \\
        Cognitive and Linguistic Sciences \\
        Brown University \\
        {\sf Mark\_Johnson@Brown.edu}
        \And
        Stefan Riezler \\
	\qquad Institut f\"ur Maschinelle Sprachverarbeitung \qquad \\
        Universit\"at Stuttgart \\
        {\sf riezler@ims.uni-stuttgart.de}}
 
\renewcommand{\Pr}{{\rm P}}
\newcommand{\Prhat}{{\widehat{\Pr}}}
\renewcommand{\L}{{\rm L}}
\newcommand{\PL}{{\rm PL}}
\newcommand{\E}{{\rm E}}
\newcommand{\V}[1]{#1}    
\newcommand{\lambdavec}{\V{\lambda}}
\newcommand{\fvec}{\V{f}}
\newcommand{\omegavec}{\tilde{\omega}}

\newcommand{\Rm}{{\R}^m}
\newcommand{\PRED}{\mbox{\sc pred}}
\newcommand{\SUBJ}{\mbox{\sc subj}}
\newcommand{\OBJ}{\mbox{\sc obj}}
\newcommand{\Johnson}{Johnson et al.~\shortcite{Johnson99c}}

\newcommand{\R}{{\rm R\hspace*{-1.65ex}\rule{0.12ex}%
       {1.55ex}\hspace*{1.65ex}}}

\begin{document}

\maketitle

\begin{abstract}
This paper describes a method for estimating conditional 
probability distributions over the parses of ``unification-based''
grammars which can utilize auxiliary distributions that are estimated
by other means.  We show how this can be used to incorporate
information about lexical selectional preferences gathered from other
sources into Stochastic ``Unification-based'' Grammars (SUBGs).  While
we apply this estimator to a Stochastic Lexical-Functional Grammar,
the method is general, and should be applicable to stochastic versions
of HPSGs, categorial grammars and transformational grammars.
\end{abstract}

\section{Introduction}
``Unification-based'' Grammars (UBGs) can capture a wide variety of
linguistically important syntactic and semantic constraints.  However,
because these constraints can be non-local or context-sensitive,
developing stochastic versions of UBGs and associated estimation
procedures is not as straight-forward as it is for, e.g., PCFGs.
Recent work has shown how to define probability distributions over the
parses of UBGs \cite{Abney97} and efficiently estimate and use
conditional probabilities for parsing \cite{Johnson99c}.  Like most
other practical stochastic grammar estimation procedures, this latter
estimation procedure requires a parsed training corpus.  

Unfortunately, large parsed UBG corpora are not yet available.  This
restricts the kinds of models one can realistically expect to be able
to estimate.  For example, a model incorporating lexical
selectional preferences of the kind described below might have tens or
hundreds of thousands of parameters, which one could not reasonably
attempt to estimate from a corpus with on the order of a thousand
clauses.  However, statistical models of lexical selectional
preferences can be estimated from very large corpora based on simpler
syntactic structures, e.g., those produced by a shallow parser.  While
there is undoubtedly disagreement between these simple syntactic
structures and the syntactic structures produced by the UBG, one might
hope that they are close enough for lexical information gathered from
the simpler syntactic structures to be of use in defining a probability
distribution over the UBG's structures.   

In the estimation procedure described here, we call the probability
distribution estimated from the larger, simpler corpus an {\em
auxiliary distribution}.  Our treatment of auxiliary distributions is
inspired by the treatment of reference distributions in
Jelinek's~\shortcite{Jelinek97} presentation of Maximum Entropy
estimation, but in our estimation procedure we simply regard the
logarithm of each auxiliary distribution as another (real-valued)
feature.  Despite its simplicity, our approach seems to offer several
advantages over the reference distribution approach.  First, it is
straight-forward to utilize several auxiliary distributions
simultaneously: each is treated as a distinct feature.  Second, each
auxiliary distribution is associated with a parameter which scales its
contribution to the final distribution.  In applications such as ours
where the auxiliary distribution may be of questionable relevance to
the distribution we are trying to estimate, it seems reasonable to
permit the estimation procedure to discount or even ignore the
auxiliary distribution.  Finally, note that neither Jelinek's nor our
estimation procedures require that an auxiliary or reference
distribution $Q$ be a probability distribution; i.e., it is not necessary
that $Q(\Omega) = 1$, where $\Omega$ is the set of well-formed linguistic
structures.

The rest of this paper is structured as follows.
Section~\ref{s:review} reviews how exponential models can be defined
over the parses of UBGs, gives a brief description of Stochastic
Lexical-Functional Grammar, and reviews why maximum pseudo-likelihood estimation
is both feasible and sufficient of parsing purposes.
Section~\ref{s:derivation} presents our new estimator, and shows how
it is related to the minimization of the Kullback-Leibler divergence
between the conditional estimated and auxiliary distributions.
Section~\ref{s:lexical} describes the
auxiliary distribution used in our experiments, and
section~\ref{s:results} presents the results of those experiments.

\section{Stochastic Unification-based Grammars} \label{s:review}
Most of the classes of probabilistic language models used in
computational linguistic are exponential families.  That is, the
probability $\Pr(\omega)$ of a well-formed syntactic structure
$\omega\in\Omega$ is defined by a function of the form
\begin{eqnarray}
 \Pr_\lambdavec(\omega) & = &  {Q(\omega) \over Z_\lambdavec}\;
                                  e ^{\lambdavec \cdot \fvec(\omega)}
   \label{e:pr1}
\end{eqnarray}
where $\fvec(\omega) \in \Rm$ is a vector of {\em feature values},
$\lambdavec \in \Rm$ is a vector of adjustable {\em feature parameters},
$Q$ is a function of $\omega$ (which Jelinek~\shortcite{Jelinek97} calls a
{\em reference distribution} when it is not an indicator function),
and $Z_\lambdavec = \int_\Omega Q(\omega) e ^{\lambdavec \cdot
\fvec(\omega)} d\omega$ is a normalization factor called the
partition function.  (Note that a feature here is just a real-valued
function of a syntactic structure $\omega$; to avoid confusion we use
the term ``attribute'' to refer to a feature in a feature structure).
If $Q(\omega) = 1$ then the class of exponential distributions is
precisely the class of distributions with maximum entropy satisfying
the constraint that the expected values of the features is a certain
specified value (e.g., a value estimated from training data), so
exponential models are sometimes also called ``Maximum Entropy''
models.

For example, the class of distributions obtained by varying the
parameters of a PCFG is an exponential family.  In a PCFG each rule or
production is associated with a feature, so $m$ is the number of rules
and the $j$th feature value $f_j(\omega)$ is the number of times the
$j$ rule is used in the derivation of the tree $\omega\in\Omega$.  Simple
manipulations show that $\Pr_\lambdavec(\omega)$ is equivalent to the
PCFG distribution if $\lambda_j = \log p_j$, where $p_j$ is the rule
emission probability, and $Q(\omega) = Z_\lambdavec = 1$.

If the features satisfy suitable Markovian independence constraints,
estimation from fully observed training data is straight-forward.  For
example, because the rule features of a PCFG meet ``context-free''
Markovian independence conditions, the well-known ``relative
frequency'' estimator for PCFGs both maximizes the likelihood of the
training data (and hence is asymptotically consistent and efficient)
and minimizes the Kullback-Leibler divergence between training and
estimated distributions.

However, the situation changes dramatically if we enforce non-local or
context-sensitive constraints on linguistic structures of the kind
that can be expressed by a UBG.  As Abney~\shortcite{Abney97} showed,
under these circumstances the relative frequency estimator is in
general inconsistent, even if one restricts attention to rule
features.  Consequently, maximum likelihood estimation is much more
complicated, as discussed in section~\ref{s:esubg}.  Moreover, while
rule features are natural for PCFGs given their context-free
independence properties, there is no particular reason to use only
rule features in Stochastic UBGs (SUBGs).  Thus an SUBG is a triple
$\langle G, f, \lambda \rangle$, where $G$ is a UBG which generates a
set of well-formed linguistic structures $\Omega$, and $f$ and
$\lambda$ are vectors of feature functions and feature parameters as
above.  The probability of a structure $\omega\in\Omega$ is given by
(\ref{e:pr1}) with $Q(\omega) = 1$.  Given a base UBG, there are
usually infinitely many different ways of selecting the features $f$
to make a SUBG, and each of these makes an empirical claim about the
class of possible distributions of structures.

\subsection{Stochastic Lexical Functional Grammar} \label{s:slfg}
Stochastic Lexical-Functional Grammar (SLFG) is a stochastic extension
of Lexical-Functional Grammar (LFG), a UBG formalism developed by
Kaplan and Bresnan~\shortcite{Kaplan82}.  Given a base LFG, an SLFG is
constructed by defining features which identify salient constructions
in a linguistic structure (in LFG this is a c-structure/f-structure
pair and its associated mapping; see Kaplan~\shortcite{Kaplan95a}).
Apart from the auxiliary distributions, we based our features on those
used in Johnson et~al.~\shortcite{Johnson99c}, which should be
consulted for further details.  Most of these feature values range
over the natural numbers, counting the number of times that a
particular construction appears in a linguistic structure.  For
example, adjunct and argument features count the number of adjunct and
argument attachments, permitting SLFG to capture a
general argument attachment preference, while more specialized
features count the number of attachments to each grammatical function
(e.g., {\sc SUBJ, OBJ, COMP}, etc.).  The flexibility of features in
stochastic UBGs permits us to include features for relatively complex
constructions, such as date expressions (it seems that date
interpretations, if possible, are usually preferred), right-branching
constituent structures (usually preferred) and non-parallel coordinate
structures (usually dispreferred).  Johnson et al.~remark that they
would have liked to have included features for lexical selectional
preferences.  While such features are perfectly acceptable in a SLFG,
they felt that their corpora were so small that the large number of
lexical dependency parameters could not be accurately estimated.  The
present paper proposes a method to address this by using an auxiliary
distribution estimated from a corpus large enough to (hopefully)
provide reliable estimates for these parameters.

\subsection{Estimating stochastic unification-based grammars} \label{s:esubg}
Suppose $\omegavec = \omega_1, \ldots, \omega_n$ is a corpus of $n$ syntactic
structures.  Letting $f_j(\omegavec) = \sum_{i=1}^n f_j(\omega_i)$ and
assuming each $\omega_i\in\Omega$, the
likelihood of the corpus $\L_\lambdavec(\omegavec)$ is:
\begin{eqnarray}
 \L_\lambdavec(\omegavec) & = & \prod_{i=1}^n \Pr_\lambdavec(\omega_i) \nonumber \\
 & = &  e^{\lambdavec\cdot f(\omegavec)} \; Z_\lambdavec^{-n}  \label{e:l}  \\
 { \partial \over \partial \lambda_j} \log \L_\lambdavec(\omegavec) & = &
   f_j(\omegavec) - n \E_\lambdavec(f_j) \label{e:dllh}
\end{eqnarray}
where $\E_\lambdavec(f_j)$ is the expected value of $f_j$ under the
distribution $\Pr_\lambdavec$.  The maximum likelihood estimates are
the $\lambda$ which maximize (\ref{e:l}), or equivalently, which make
(\ref{e:dllh}) zero, but as \Johnson\ explain, there seems to be no
practical way of computing these for realistic SUBGs since evaluating
(\ref{e:l}) and its derivatives (\ref{e:dllh}) involves integrating
over all syntactic structures $\Omega$.

However, Johnson et al.~observe that parsing applications require only
the {\em conditional} probability distribution
$\Pr_\lambdavec(\omega|y)$, where $y$ is the terminal string or {\em
yield} being parsed, and that this can be estimated by maximizing the
{\em pseudo-likelihood} of the corpus $\PL_\lambdavec(\omegavec)$:
\begin{eqnarray}
 \PL_\lambdavec(\omegavec) & = & \prod_{i=1}^n \Pr_\lambdavec(\omega_i|y_i) \nonumber \\
 & = & e^{\lambdavec\cdot f(\omegavec)} \; \prod_{i=1}^n Z_\lambdavec^{-1}(y_i)
              \label{e:pl} 
\end{eqnarray}
In (\ref{e:pl}), $y_i$ is the yield of $\omega_i$ and
\[
 Z_\lambdavec(y_i) = \int_{\Omega(y_i)} e^{\lambdavec\cdot f(\omega)}
d\omega,
\]
where $\Omega(y_i)$ is the set of all syntactic structures
in $\Omega$ with yield $y_i$ (i.e., all parses of $y_i$ generated by
the base UBG).  It turns out that calculating the pseudo-likelihood of
a corpus only involves integrations over the sets of parses of its
yields $\Omega(y_i)$, which is feasible for many interesting UBGs.
Moreover, the maximum pseudo-likelihood estimator is asymptotically
consistent for the conditional distribution $\Pr(\omega|y)$.  For the
reasons explained in \Johnson\ we actually estimate $\lambdavec$ by
maximizing a regularized version of the log pseudo-likelihood
(\ref{e:rlpl}), where $\sigma_j$ is 7 times the maximum value of $f_j$
found in the training corpus:
\begin{equation}
 \log \PL_\lambdavec(\omegavec) - \sum_{j=1}^m {\lambda_j^2 \over 2 \sigma_j^2}
 \label{e:rlpl}
\end{equation}
See \Johnson\ for details of the calculation of this quantity and its derivatives,
and the conjugate gradient routine used to calculate the $\lambda$ which
maximize the regularized log pseudo-likelihood of the training corpus.

\section{Auxiliary distributions} \label{s:derivation}
We modify the estimation problem presented in section~\ref{s:esubg} by
assuming that in addition to the corpus $\omegavec$ and the $m$
feature functions $f$ we are given $k$ auxiliary distributions $Q_1,
\ldots, Q_k$ whose support includes $\Omega$ that we suspect may be
related to the joint distribution $\Pr(\omega)$ or conditional
distribution $\Pr(\omega|y)$ that we wish to estimate.  We do not
require that the $Q_j$ be probability distributions, i.e., it is not
necessary that $\int_\Omega Q_j(\omega) d\omega = 1$, but we do
require that they are strictly positive (i.e., $Q_j(\omega) > 0,
\forall \omega\in\Omega$).  We define $k$ new features $f_{m+1},
\ldots, f_{m+k}$ where $f_{m+j}(\omega) =
\log Q_j(\omega)$, which we call {\em auxiliary features}.  The $m+k$
parameters associated with the resulting $m+k$ features can be
estimated using any method for estimating the parameters of an
exponential family with real-valued features (in our experiments we
used the pseudo-likelihood estimation procedure reviewed in
section~\ref{s:esubg}).  Such a procedure estimates parameters
$\lambda_{m+1}, \ldots, \lambda_{m+k}$ associated with the auxiliary
features, so the estimated distributions take the form (\ref{e:apr})
(for simplicity we only discuss joint distributions here, but the
treatment of conditional distributions is parallel).
\begin{eqnarray}
\Pr_\lambdavec(\omega) & = & { 
       \prod_{j=1}^k Q_j(\omega)^{\lambda_{m+j}} \over Z_\lambdavec}
 \; e ^ {\sum_{j=1}^m \lambda_j f_j(\omega) }. \label{e:apr}
\end{eqnarray}
Note that the auxiliary distributions $Q_j$ are treated as fixed
distributions for the purposes of this estimation, even though each
$Q_j$ may itself be a complex model obtained via a previous
estimation process.  Comparing (\ref{e:apr}) with (\ref{e:pr1}) on
page~\pageref{e:pr1}, we see that the two equations become identical
if the reference distribution $Q$ in (\ref{e:pr1}) is replaced by a
geometric mixture of the auxiliary distributions $Q_j$, i.e., if:
\begin{eqnarray*}
 Q(\omega) & =  & 
                   \prod_{j=1}^k Q_j(\omega)^{\lambda_{m+j}}.
\end{eqnarray*}
The parameter associated with an auxiliary feature represents
the weight of that feature in the mixture.  If a parameter
$\lambda_{m+j} = 1$ then the corresponding auxiliary feature $Q_j$ is
equivalent to a reference distribution in Jelinek's sense, while if
$\lambda_{m+j} = 0$ then $Q_j$ is effectively ignored.  Thus our approach
can be regarded as a smoothed version Jelinek's reference distribution
approach, generalized to permit multiple auxiliary distributions.

\section{Lexical selectional preferences} \label{s:lexical}
The auxiliary distribution we used here is based on the probabilistic model
of lexical selectional preferences described in Rooth
et~al.~\shortcite{Rooth99}.  An existing broad-coverage parser was
used to find shallow parses (compared to the LFG parses) for the 117
million word British National Corpus \cite{Carroll98}.  We based our
auxiliary distribution on 3.7 million $\langle g, r, a\rangle$ tuples
(belonging to 600,000 types) we extracted these parses, where $g$ is a
lexical governor (for the shallow parses, $g$ is either a verb or a
preposition), $a$ is the head of one of its NP arguments and $r$ is
the the grammatical relationship between the governor and argument (in
the shallow parses $r$ is always $\OBJ$ for prepositional governors,
and $r$ is either $\SUBJ$ or $\OBJ$ for verbal governors).

In order to avoid sparse data problems we smoothed this distribution
over tuples as described in \cite{Rooth99}.  We assume that
governor-relation pairs $\langle g, r\rangle$ and arguments $a$ are
independently generated from 25 hidden classes $C$, i.e.:
\begin{eqnarray*}
 \Prhat(\langle g,r,a \rangle) & = & \sum_{c \in C} 
        \Pr_e(\langle g,r \rangle | c) \Prhat_e(a|c) \Pr_e(c)
\end{eqnarray*}
where the distributions $\Pr_e$ are estimated from the training tuples
using the Expectation-Maximization algorithm.  While the hidden
classes are not given any prior interpretation they often cluster
semantically coherent predicates and arguments, as shown in
Figure~\ref{f:sem}.  The smoothing power of a clustering model such as
this can be calculated explicitly as the percentage of possible tuples
which are assigned a non-zero probability.  For the 25-class model we
get a smoothing power of $99\%$, compared to only $1.7\%$ using the
empirical distribution of the training data.

\begin{figure*}
\begin{center}
\setlength{\tabcolsep}{2pt}
{\tiny
\begin{tabular}
{|l|r|rrrrrrrrrrrrrrrrrrrrrrrrrrrrrr|} \hline
\begin{tabular}{c}
     {\bf Class 16} \\
                                 \\
     PROB 0.0340\\
                                 \\
\end{tabular}
&  & \rotate[l]{0.0158} 
 & \rotate[l]{0.0121} 
 & \rotate[l]{0.0081} 
 & \rotate[l]{0.0079} 
 & \rotate[l]{0.0075} 
 & \rotate[l]{0.0058} 
 & \rotate[l]{0.0055} 
 & \rotate[l]{0.0055} 
 & \rotate[l]{0.0052} 
 & \rotate[l]{0.0050} 
 & \rotate[l]{0.0049} 
 & \rotate[l]{0.0048} 
 & \rotate[l]{0.0047} 
 & \rotate[l]{0.0047} 
 & \rotate[l]{0.0046} 
 & \rotate[l]{0.0046} 
 & \rotate[l]{0.0045} 
 & \rotate[l]{0.0045} 
 & \rotate[l]{0.0041} 
 & \rotate[l]{0.0041} 
 & \rotate[l]{0.0039} 
 & \rotate[l]{0.0039} 
 & \rotate[l]{0.0038} 
 & \rotate[l]{0.0038} 
 & \rotate[l]{0.0037} 
 & \rotate[l]{0.0036} 
 & \rotate[l]{0.0036} 
 & \rotate[l]{0.0036} 
 & \rotate[l]{0.0035} 
 & \rotate[l]{0.0035} 
 \\ 
 \hline
&  & \rotate[l]{spokesman} 
 & \rotate[l]{we} 
 & \rotate[l]{people} 
 & \rotate[l]{mother} 
 & \rotate[l]{doctor} 
 & \rotate[l]{police} 
 & \rotate[l]{woman} 
 & \rotate[l]{father} 
 & \rotate[l]{director} 
 & \rotate[l]{night} 
 & \rotate[l]{someone} 
 & \rotate[l]{report} 
 & \rotate[l]{officer} 
 & \rotate[l]{john} 
 & \rotate[l]{girl} 
 & \rotate[l]{official} 
 & \rotate[l]{ruth} 
 & \rotate[l]{voice} 
 & \rotate[l]{stephen} 
 & \rotate[l]{company} 
 & \rotate[l]{god} 
 & \rotate[l]{chairman} 
 & \rotate[l]{no-one} 
 & \rotate[l]{man} 
 & \rotate[l]{who} 
 & \rotate[l]{edward} 
 & \rotate[l]{mum} 
 & \rotate[l]{nobody} 
 & \rotate[l]{everyone} 
 & \rotate[l]{peter} 
 \\ 
 \hline
0.3183 & say:s & $\bullet$ & $\bullet$ & $\bullet$ & $\bullet$ & $\bullet$ & $\bullet$ & $\bullet$ & $\bullet$ & $\bullet$ & $\bullet$ & $\bullet$ & $\bullet$ & $\bullet$ & $\bullet$ & $\bullet$ & $\bullet$ & $\bullet$ & $\bullet$ & $\bullet$ & $\bullet$ & $\bullet$ & $\bullet$ & $\bullet$ &  & $\bullet$ & $\bullet$ & $\bullet$ & $\bullet$ & $\bullet$ & $\bullet$\\
0.0405 & say:o & $\bullet$ & $\bullet$ & $\bullet$ & $\bullet$ & $\bullet$ & $\bullet$ & $\bullet$ & $\bullet$ & $\bullet$ & $\bullet$ & $\bullet$ & $\bullet$ & $\bullet$ & $\bullet$ & $\bullet$ & $\bullet$ & $\bullet$ & $\bullet$ & $\bullet$ & $\bullet$ & $\bullet$ & $\bullet$ &  &  & $\bullet$ & $\bullet$ & $\bullet$ &  & $\bullet$ & $\bullet$\\
0.0345 & ask:s & $\bullet$ & $\bullet$ & $\bullet$ & $\bullet$ & $\bullet$ & $\bullet$ & $\bullet$ & $\bullet$ & $\bullet$ & $\bullet$ & $\bullet$ & $\bullet$ & $\bullet$ & $\bullet$ & $\bullet$ & $\bullet$ & $\bullet$ & $\bullet$ & $\bullet$ & $\bullet$ & $\bullet$ & $\bullet$ & $\bullet$ & $\bullet$ & $\bullet$ & $\bullet$ & $\bullet$ & $\bullet$ & $\bullet$ & $\bullet$\\
0.0276 & tell:s & $\bullet$ & $\bullet$ & $\bullet$ & $\bullet$ & $\bullet$ & $\bullet$ & $\bullet$ & $\bullet$ & $\bullet$ & $\bullet$ & $\bullet$ & $\bullet$ & $\bullet$ & $\bullet$ & $\bullet$ & $\bullet$ & $\bullet$ & $\bullet$ & $\bullet$ & $\bullet$ & $\bullet$ & $\bullet$ & $\bullet$ & $\bullet$ & $\bullet$ & $\bullet$ & $\bullet$ & $\bullet$ & $\bullet$ & $\bullet$\\
0.0214 & be:s & $\bullet$ &  & $\bullet$ & $\bullet$ & $\bullet$ & $\bullet$ & $\bullet$ & $\bullet$ & $\bullet$ & $\bullet$ & $\bullet$ &  & $\bullet$ & $\bullet$ & $\bullet$ & $\bullet$ & $\bullet$ &  & $\bullet$ & $\bullet$ & $\bullet$ & $\bullet$ & $\bullet$ &  &  & $\bullet$ & $\bullet$ & $\bullet$ & $\bullet$ & $\bullet$\\
0.0193 & know:s & $\bullet$ &  & $\bullet$ & $\bullet$ & $\bullet$ & $\bullet$ & $\bullet$ & $\bullet$ & $\bullet$ &  & $\bullet$ &  & $\bullet$ & $\bullet$ & $\bullet$ & $\bullet$ & $\bullet$ &  & $\bullet$ & $\bullet$ & $\bullet$ &  & $\bullet$ & $\bullet$ &  & $\bullet$ & $\bullet$ & $\bullet$ & $\bullet$ & $\bullet$\\
0.0147 & have:s &  &  &  &  & $\bullet$ & $\bullet$ & $\bullet$ & $\bullet$ & $\bullet$ & $\bullet$ & $\bullet$ & $\bullet$ & $\bullet$ & $\bullet$ & $\bullet$ & $\bullet$ & $\bullet$ & $\bullet$ & $\bullet$ &  & $\bullet$ & $\bullet$ & $\bullet$ & $\bullet$ &  & $\bullet$ & $\bullet$ & $\bullet$ & $\bullet$ & $\bullet$\\
0.0144 & nod:s &  & $\bullet$ & $\bullet$ & $\bullet$ & $\bullet$ &  & $\bullet$ & $\bullet$ & $\bullet$ &  &  &  & $\bullet$ & $\bullet$ & $\bullet$ &  & $\bullet$ &  & $\bullet$ &  &  & $\bullet$ &  & $\bullet$ &  & $\bullet$ & $\bullet$ &  & $\bullet$ & $\bullet$\\
0.0137 & think:s &  & $\bullet$ & $\bullet$ & $\bullet$ & $\bullet$ & $\bullet$ & $\bullet$ & $\bullet$ & $\bullet$ & $\bullet$ & $\bullet$ &  & $\bullet$ & $\bullet$ & $\bullet$ & $\bullet$ & $\bullet$ &  & $\bullet$ &  & $\bullet$ & $\bullet$ & $\bullet$ & $\bullet$ & $\bullet$ & $\bullet$ & $\bullet$ & $\bullet$ & $\bullet$ & $\bullet$\\
0.0130 & shake:s &  & $\bullet$ & $\bullet$ & $\bullet$ & $\bullet$ &  & $\bullet$ & $\bullet$ &  &  &  &  & $\bullet$ & $\bullet$ & $\bullet$ &  & $\bullet$ & $\bullet$ & $\bullet$ &  &  &  &  & $\bullet$ &  & $\bullet$ & $\bullet$ & $\bullet$ & $\bullet$ & $\bullet$\\
0.0128 & take:s & $\bullet$ &  & $\bullet$ & $\bullet$ & $\bullet$ & $\bullet$ & $\bullet$ & $\bullet$ & $\bullet$ & $\bullet$ & $\bullet$ & $\bullet$ & $\bullet$ & $\bullet$ & $\bullet$ & $\bullet$ & $\bullet$ & $\bullet$ & $\bullet$ & $\bullet$ & $\bullet$ & $\bullet$ & $\bullet$ & $\bullet$ & $\bullet$ & $\bullet$ & $\bullet$ & $\bullet$ & $\bullet$ & $\bullet$\\
0.0104 & reply:s &  & $\bullet$ & $\bullet$ & $\bullet$ & $\bullet$ &  & $\bullet$ & $\bullet$ & $\bullet$ &  &  &  & $\bullet$ & $\bullet$ & $\bullet$ & $\bullet$ & $\bullet$ & $\bullet$ & $\bullet$ &  & $\bullet$ & $\bullet$ & $\bullet$ & $\bullet$ & $\bullet$ & $\bullet$ & $\bullet$ &  &  & $\bullet$\\
0.0096 & smile:s &  & $\bullet$ & $\bullet$ & $\bullet$ & $\bullet$ &  & $\bullet$ & $\bullet$ &  &  & $\bullet$ &  & $\bullet$ & $\bullet$ & $\bullet$ &  & $\bullet$ &  & $\bullet$ &  &  & $\bullet$ &  & $\bullet$ &  & $\bullet$ & $\bullet$ & $\bullet$ & $\bullet$ & $\bullet$\\
0.0094 & do:s &  &  &  & $\bullet$ & $\bullet$ & $\bullet$ & $\bullet$ & $\bullet$ & $\bullet$ & $\bullet$ & $\bullet$ & $\bullet$ & $\bullet$ & $\bullet$ & $\bullet$ & $\bullet$ & $\bullet$ & $\bullet$ & $\bullet$ & $\bullet$ & $\bullet$ & $\bullet$ & $\bullet$ & $\bullet$ & $\bullet$ & $\bullet$ & $\bullet$ & $\bullet$ & $\bullet$ & $\bullet$\\
0.0094 & laugh:s &  & $\bullet$ & $\bullet$ & $\bullet$ & $\bullet$ &  & $\bullet$ & $\bullet$ & $\bullet$ & $\bullet$ & $\bullet$ &  &  & $\bullet$ & $\bullet$ &  & $\bullet$ & $\bullet$ & $\bullet$ & $\bullet$ &  &  & $\bullet$ & $\bullet$ & $\bullet$ & $\bullet$ & $\bullet$ & $\bullet$ & $\bullet$ & $\bullet$\\
0.0089 & tell:o &  &  & $\bullet$ & $\bullet$ & $\bullet$ & $\bullet$ & $\bullet$ & $\bullet$ & $\bullet$ & $\bullet$ & $\bullet$ & $\bullet$ & $\bullet$ & $\bullet$ & $\bullet$ & $\bullet$ & $\bullet$ &  & $\bullet$ & $\bullet$ & $\bullet$ &  & $\bullet$ & $\bullet$ & $\bullet$ & $\bullet$ & $\bullet$ & $\bullet$ & $\bullet$ & $\bullet$\\
0.0084 & saw:s &  & $\bullet$ & $\bullet$ & $\bullet$ & $\bullet$ & $\bullet$ & $\bullet$ & $\bullet$ &  & $\bullet$ & $\bullet$ &  & $\bullet$ & $\bullet$ & $\bullet$ & $\bullet$ & $\bullet$ &  & $\bullet$ & $\bullet$ & $\bullet$ & $\bullet$ & $\bullet$ & $\bullet$ & $\bullet$ & $\bullet$ & $\bullet$ & $\bullet$ & $\bullet$ & $\bullet$\\
0.0082 & add:s & $\bullet$ &  & $\bullet$ & $\bullet$ & $\bullet$ & $\bullet$ & $\bullet$ &  & $\bullet$ & $\bullet$ &  & $\bullet$ & $\bullet$ & $\bullet$ &  & $\bullet$ & $\bullet$ & $\bullet$ & $\bullet$ & $\bullet$ &  & $\bullet$ &  & $\bullet$ & $\bullet$ & $\bullet$ & $\bullet$ &  &  & $\bullet$\\
0.0078 & feel:s &  & $\bullet$ & $\bullet$ & $\bullet$ & $\bullet$ & $\bullet$ & $\bullet$ & $\bullet$ & $\bullet$ &  & $\bullet$ &  & $\bullet$ & $\bullet$ & $\bullet$ & $\bullet$ & $\bullet$ &  & $\bullet$ & $\bullet$ & $\bullet$ &  & $\bullet$ & $\bullet$ & $\bullet$ & $\bullet$ & $\bullet$ & $\bullet$ & $\bullet$ & $\bullet$\\
0.0071 & make:s & $\bullet$ &  & $\bullet$ & $\bullet$ & $\bullet$ & $\bullet$ & $\bullet$ & $\bullet$ & $\bullet$ & $\bullet$ & $\bullet$ & $\bullet$ & $\bullet$ & $\bullet$ & $\bullet$ & $\bullet$ & $\bullet$ & $\bullet$ & $\bullet$ & $\bullet$ & $\bullet$ & $\bullet$ & $\bullet$ & $\bullet$ & $\bullet$ & $\bullet$ & $\bullet$ & $\bullet$ & $\bullet$ & $\bullet$\\
0.0070 & give:s & $\bullet$ & $\bullet$ & $\bullet$ & $\bullet$ & $\bullet$ & $\bullet$ & $\bullet$ & $\bullet$ & $\bullet$ & $\bullet$ & $\bullet$ & $\bullet$ & $\bullet$ & $\bullet$ & $\bullet$ & $\bullet$ & $\bullet$ & $\bullet$ & $\bullet$ & $\bullet$ & $\bullet$ & $\bullet$ & $\bullet$ & $\bullet$ & $\bullet$ & $\bullet$ & $\bullet$ & $\bullet$ & $\bullet$ & $\bullet$\\
0.0067 & ask:o & $\bullet$ &  & $\bullet$ & $\bullet$ & $\bullet$ & $\bullet$ & $\bullet$ & $\bullet$ & $\bullet$ & $\bullet$ & $\bullet$ &  & $\bullet$ & $\bullet$ & $\bullet$ &  & $\bullet$ &  & $\bullet$ & $\bullet$ & $\bullet$ & $\bullet$ & $\bullet$ & $\bullet$ & $\bullet$ &  & $\bullet$ & $\bullet$ & $\bullet$ & $\bullet$\\
0.0066 & shrug:s &  & $\bullet$ & $\bullet$ & $\bullet$ & $\bullet$ &  & $\bullet$ & $\bullet$ & $\bullet$ &  &  &  & $\bullet$ &  & $\bullet$ &  &  &  & $\bullet$ &  &  &  &  & $\bullet$ &  & $\bullet$ &  &  &  & $\bullet$\\
0.0061 & explain:s & $\bullet$ & $\bullet$ & $\bullet$ & $\bullet$ & $\bullet$ &  & $\bullet$ & $\bullet$ & $\bullet$ &  & $\bullet$ & $\bullet$ & $\bullet$ & $\bullet$ &  & $\bullet$ & $\bullet$ & $\bullet$ & $\bullet$ & $\bullet$ & $\bullet$ & $\bullet$ & $\bullet$ & $\bullet$ & $\bullet$ & $\bullet$ & $\bullet$ & $\bullet$ &  & $\bullet$\\
0.0051 & like:s &  & $\bullet$ & $\bullet$ & $\bullet$ &  &  & $\bullet$ & $\bullet$ & $\bullet$ &  & $\bullet$ &  & $\bullet$ & $\bullet$ & $\bullet$ &  &  &  & $\bullet$ & $\bullet$ &  & $\bullet$ & $\bullet$ & $\bullet$ & $\bullet$ & $\bullet$ & $\bullet$ & $\bullet$ & $\bullet$ & $\bullet$\\
0.0050 & look:s &  &  & $\bullet$ & $\bullet$ & $\bullet$ & $\bullet$ & $\bullet$ & $\bullet$ &  &  &  & $\bullet$ &  & $\bullet$ & $\bullet$ &  & $\bullet$ &  & $\bullet$ & $\bullet$ & $\bullet$ &  & $\bullet$ & $\bullet$ & $\bullet$ & $\bullet$ & $\bullet$ & $\bullet$ & $\bullet$ & \\
0.0050 & sigh:s &  & $\bullet$ & $\bullet$ & $\bullet$ & $\bullet$ &  & $\bullet$ & $\bullet$ &  &  &  &  &  & $\bullet$ &  &  & $\bullet$ &  & $\bullet$ &  &  & $\bullet$ &  & $\bullet$ &  & $\bullet$ &  &  &  & $\bullet$\\
0.0049 & watch:s &  & $\bullet$ & $\bullet$ & $\bullet$ & $\bullet$ & $\bullet$ & $\bullet$ & $\bullet$ & $\bullet$ & $\bullet$ & $\bullet$ &  & $\bullet$ & $\bullet$ & $\bullet$ &  & $\bullet$ &  & $\bullet$ &  & $\bullet$ & $\bullet$ & $\bullet$ & $\bullet$ & $\bullet$ &  & $\bullet$ & $\bullet$ & $\bullet$ & $\bullet$\\
0.0049 & hear:s &  &  & $\bullet$ & $\bullet$ & $\bullet$ & $\bullet$ & $\bullet$ & $\bullet$ & $\bullet$ &  & $\bullet$ &  & $\bullet$ & $\bullet$ & $\bullet$ &  & $\bullet$ & $\bullet$ &  &  & $\bullet$ & $\bullet$ & $\bullet$ & $\bullet$ & $\bullet$ &  & $\bullet$ & $\bullet$ & $\bullet$ & $\bullet$\\
0.0047 & answer:s &  & $\bullet$ & $\bullet$ & $\bullet$ & $\bullet$ & $\bullet$ & $\bullet$ & $\bullet$ &  &  & $\bullet$ &  & $\bullet$ & $\bullet$ & $\bullet$ &  & $\bullet$ & $\bullet$ &  &  & $\bullet$ &  & $\bullet$ & $\bullet$ & $\bullet$ & $\bullet$ & $\bullet$ & $\bullet$ &  & \\
\hline\end{tabular}}
 \end{center}
  \caption{A depiction of the highest probability predicates and arguments
  in Class~16.  The class matrix shows at the top the 30 most probable nouns in the
   $\Pr_e(a|16)$ distribution and their probabilities, and at the left the 30
   most probable verbs and prepositions listed according to $Pr_e(\langle g,r\rangle|16)$ 
   and their probabilities. Dots in the matrix indicate that the respective pair
   was seen in the training data. Predicates with suffix $:s$ indicate
   the subject slot of an intransitive or transitive verb; the suffix
   $:o$ specifies the nouns in the corresponding row as objects of verbs 
   or prepositions. 
  \label{f:sem}}
\end{figure*}

\section{Empirical evaluation} \label{s:results}
Hadar Shemtov and Ron Kaplan at Xerox {\sc Parc} provided us with two
LFG parsed corpora called the Verbmobil corpus and the Homecentre
corpus.  These contain parse forests for each sentence (packed
according to scheme described in Maxwell and
Kaplan~\shortcite{Maxwell95b}), together with a manual annotation as
to which parse is correct.  The Verbmobil corpus contains
540~sentences relating to appointment planning, while the Homecentre
corpus contains 980~sentences from Xerox documentation on their
``homecentre'' multifunction devices.  Xerox did not provide us with
the base LFGs for intellectual property reasons, but from inspection
of the parses it seems that slightly different grammars were used with
each corpus, so we did not merge the corpora.  We chose the features
of our SLFG based solely on the basis of the Verbmobil corpus, so the
Homecentre corpus can be regarded as a held-out evaluation corpus.

We discarded the unambiguous sentences in each corpus for both
training and testing (as explained in \Johnson, pseudo-likelihood
estimation ignores unambiguous sentences), leaving us with a corpus of
324~ambiguous sentences in the Verbmobil corpus and 481~sentences in
the Homecentre corpus; these sentences had a total of 3,245~and 
3,169~parses respectively.

The (non-auxiliary) features used in were based on those described by
\Johnson.  Different numbers of features were used with the two
corpora because some of the features were generated semi-automatically
(e.g., we introduced a feature for every attribute-value pair found in
any feature structure), and ``pseudo-constant'' features (i.e.,
features whose values never differ on the parses of the same sentence)
are discarded.  We used 172~features in the SLFG for the Verbmobil
corpus and 186~features in the SLFG for the Homecentre corpus.

We used three additional auxiliary features derived from the lexical
selectional preference model described in section~\ref{s:lexical}.
These were defined in the following way.  For each governing predicate
$g$, grammatical relation $r$ and argument $a$, let $n_{\langle g, r, a\rangle}(\omega)$
be the number of times that the f-structure:
\[
 \left[ 
  \begin{array}{l}
  \PRED = g \\
  r = \left[ \PRED = a \right]
  \end{array}
 \right]
\]
appears as a subgraph of the f-structure of $\omega$, i.e., the number of
times that $a$ fills the grammatical role $r$ of $g$.  We used the lexical
model described in the last section to estimate $\Prhat(a|g,r)$, and defined
our first auxiliary feature as:
\begin{eqnarray*}
 f_l(\omega) & = & \log \Prhat(g_0) +
                   \sum_{\langle g, r, a\rangle} n_{\langle g, r, a\rangle}(\omega) 
                   \log \Prhat(a|g,r)
\end{eqnarray*}
where $g_0$ is the predicate of the root feature structure.  The
justification for this feature is that if f-structures were in fact a
tree, $f_l(\omega)$ would be the (logarithm of) a probability
distribution over them.  The auxiliary feature $f_l$ is defective in
many ways.  Because LFG f-structures are DAGs with reentrancies rather
than trees we double count certain arguments, so $f_l$ is certainly
not the logarithm of a probability distribution (which is why we
stressed that our approach does not require an auxiliary distribution to be
a distribution).  

The number of governor-argument tuples found in different parses of
the same sentence can vary markedly.  Since the conditional
probabilities $\Prhat(a|g,r)$ are usually very small, we found that
$f_l(\omega)$ was strongly related to the number of tuples found in
$\omega$, so the parse with the smaller number of tuples usually
obtains the higher $f_l$ score.  
We tried to address this by adding two additional
features.  We set $f_c(\omega)$ to be the number of tuples in $\omega$, i.e.:
\begin{eqnarray*}
 f_c(\omega) & = & \sum_{\langle g, r, a\rangle} n_{\langle g, r,
a\rangle}(\omega).
\end{eqnarray*}
Then we set $f_n(\omega) = f_l(\omega)/f_c(\omega)$, i.e., $f_n(\omega)$
is the average log probability of a lexical dependency tuple under the
auxiliary lexical distribution.  We performed our experiments with $f_l$
as the sole auxiliary distribution, and with $f_l$, $f_c$ and $f_n$ as three
auxiliary distributions.

Because our corpora were so small, we trained and tested these models
using a 10-fold cross-validation paradigm; the cumulative results are
shown in Table~\ref{t:results}.  On each fold we evaluated each model
in two ways.  The {\em correct parses measure} simply counts the
number of test sentences for which the estimated model assigns its
maximum parse probability to the correct parse, with ties broken
randomly.  The {\em pseudo-likelihood measure} is the
pseudo-likelihood of test set parses; i.e., the conditional
probability of the test parses given their yields.  We actually report
the negative log of this measure, so a smaller score corresponds to
better performance here.  The correct parses measure is most closely
related to parser performance, but the pseudo-likelihood measure is
more closely related to the quantity we are optimizing and may be more
relevant to applications where the parser has to return a certainty
factor associated with each parse.

Table~\ref{t:results} also provides the number of {\em
indistinguishable sentences} under each model.  A sentence $y$ is
indistinguishable with respect to features $f$ iff $f(\omega_c) =
f(\omega')$, where $\omega_c$ is the correct parse of $y$ and
$\omega_c \neq \omega' \in \Omega(y)$, i.e., the feature values of
correct parse of $y$ are identical to the feature values of some other
parse of $y$.  If a sentence is indistinguishable it is not possible
to assign its correct parse a (conditional) probability higher than
the (conditional) probability assigned to other parses, so all else
being equal we would expect a SUBG with with fewer indistinguishable
sentences to perform better than one with more.

\begin{table*}
\begin{center}
\begin{tabular}{cccc}
\multicolumn{4}{c}{{\bf Verbmobil corpus} (324 sentences, 172 non-auxiliary features)} \\
\bf Auxiliary features used & \bf Indistinguishable & \bf Correct & \bf - log PL \\
(none) & 9 & 180 & 401.3 \\
$f_l$ & 8 & 183 & 401.6 \\
$f_l, f_c, f_n$ & 8 & 180.5 & 404.0 \\ 
\ \\
\multicolumn{4}{c}{{\bf Homecentre corpus} (481 sentences, 186 non-auxiliary features)} \\
\bf Auxiliary features used & \bf Indistinguishable & \bf Correct & \bf - log PL \\
(none) & 45 & 283.25 & 580.6 \\
$f_l$ & 34 & 284 & 580.6 \\
$f_l, f_c, f_n$ & 34 & 285 & 582.2
\end{tabular}
\end{center}
\caption{ \label{t:results}
 The effect of adding auxiliary lexical dependency features to a SLFG.
 The auxiliary features are described in the text. The column labelled
 ``indistinguishable'' gives the number of indistinguishable sentences
 with respect to each feature set, while ``correct'' and ``-- log PL''
 give the correct parses and pseudo-likelihood measures respectively.}
\end{table*}

Adding auxiliary features reduced the already low number
of indistinguishable sentences in the Verbmobil corpus by only 11\%,
while it reduced the number of indistinguishable sentences in the
Homecentre corpus by 24\%.
This probably reflects the fact that the feature set was designed 
by inspecting only the Verbmobil corpus.

We must admit disappointment with these results.  Adding auxiliary
lexical features improves the correct parses measure only slightly,
and degrades rather than improves performance on the pseudo-likelihood
measure.  Perhaps this is due to the fact that adding auxiliary
features increases the dimensionality of the feature vector $f$, so
the pseudo-likelihood scores with different numbers of features are
not strictly comparable.

The small improvement in the correct parses measure is typical of the
improvement we might expect to achieve by adding a ``good''
non-auxiliary feature, but given the importance usually placed on
lexical dependencies in statistical models one might have expected
more improvement.  
Probably the poor performance is due in part to the fairly large differences
between the parses from which the lexical dependencies were estimated
and the parses produced by the LFG. 
LFG parses are very detailed, and many ambiguities depend on the precise
grammatical relationship holding between a predicate and its argument.
It could also be that better
performance could be achieved if the lexical dependencies were
estimated from a corpus more closely related to the actual test
corpus.  For example, the verb {\em feed} in the Homecentre corpus is
used in the sense of ``insert (paper into printer)'', which hardly
seems to be a prototypical usage.

Note that overall system performance is quite good; taking the
unambiguous sentences into account the combined LFG parser and
statistical model finds the correct parse for 73\% of the Verbmobil
test sentences and 80\% of the Homecentre test sentences.  On just the
ambiguous sentences, our system selects the correct parse for 56\% 
of the Verbmobil test sentences and 59\% of the Homecentre test 
sentences.

\section{Conclusion}
This paper has presented a method for incorporating auxiliary
distributional information gathered by other means possibly from other
corpora into a Stochastic ``Unification-based'' Grammar (SUBG).  This
permits one to incorporate dependencies into a SUBG which probably
cannot be estimated directly from the small UBG parsed corpora
available today.  It has the virtue that it can incorporate several
auxiliary distributions simultaneously, and because it associates each
auxiliary distribution with its own ``weight'' parameter, it can scale
the contributions of each auxiliary distribution toward the final
estimated distribution, or even ignore it entirely.  We have applied
this to incorporate lexical selectional preference information into a
Stochastic Lexical-Functional Grammar, but the technique generalizes
to stochastic versions of HPSGs, categorial grammars and
transformational grammars.  An obvious extension of this work, which we
hope will be persued in the future,  is to apply these
techniques in broad-coverage feature-based TAG parsers.

\bibliographystyle{acl}

\bibliography{mj}

\end{document}